\documentclass[12pt]{article}
\usepackage{amsmath}
\usepackage{graphicx}
\usepackage{natbib}
\usepackage{url} 
\usepackage{amsmath,amssymb,amsthm}
\usepackage{multirow,multicol}
\usepackage{latexsym}
\usepackage{bm}
\usepackage{tikz}
\usepackage{dsfont}

\usepackage{amsmath,amssymb,amsthm}
\usepackage{multirow,multicol}
\usepackage{latexsym}
\usepackage{bm}
\usepackage{tikz}
\usepackage{dsfont}
\usepackage{rotating}
\usepackage{mathabx}
\RequirePackage{algorithm}
\RequirePackage{algorithmic}
\usepackage{array}
\newcolumntype{H}{>{\setbox0=\hbox\bgroup}c<{\egroup}@{}}
\newcolumntype{Z}{>{\setbox0=\hbox\bgroup}c<{\egroup}@{\hspace*{-\tabcolsep}}}
\usetikzlibrary{arrows.meta, positioning}
\newcommand{\blind}{0}
\newcommand{\inctabfig}{1}

\DeclareMathOperator*{\argmax}{arg\,max}

\newcommand{\bx}{\bm{x}}

\newcommand{\bz}{\bm{z}}

\newcommand{\bX}{\bm{X}}

\newcommand{\bZ}{\bm{Z}}

\newcommand{\var}{\mathrm{var}}

\newcommand{\mbE}{\mathbb{E}}
\newcommand{\mbP}{\mathbb{P}}
\newcommand{\mcA}{\mathcal{A}}
\newcommand{\mcI}{\mathcal{I}}
\newcommand{\mbQ}{\mathbb{Q}}
\newcommand{\E}{\mathbb{E}}

\newcommand{\ind}[1]{\mathds{I}{(#1)}}

\newtheorem{theorem}{Theorem}
\newtheorem{remark}{Remark}

\newtheorem{lemma}{Lemma}

\begin{document}

\def\spacingset#1{\renewcommand{\baselinestretch}%
{#1}\small\normalsize} \spacingset{1}


\if0\blind
{
\title{Nonparametric Feature Selection by Random Forests and Deep Neural Networks}
  \author{Xiaojun Mao\\
   School of Mathematical Sciences, \\Shanghai Jiao Tong University,  Shanghai, China\\
     \\
    Liuhua Peng \\
    School of Mathematics and Statistics, \\The University of Melbourne, Victoria, Australia\\
and\\
Zhonglei Wang\thanks{Correspondence to wangzl@xmu.edu.cn.}\\
 Wang Yanan Institute for Studies in Economics, \\Xiamen University, 
	Xiamen, Fujian, China
}
\date{}
  \maketitle
} \fi

\if1\blind
{
  \bigskip
  \bigskip
  \bigskip
  \begin{center}
    {\LARGE\bf Supplement for ``Doubly Efficient Random Forest''}
\end{center}
  \medskip
} \fi

\bigskip
\spacingset{1.5} 

\begin{abstract}
	Random forests are a widely used machine learning algorithm, but their computational efficiency is undermined when applied to large-scale datasets with numerous instances and useless features. Herein, we propose a nonparametric feature selection algorithm that incorporates random forests and deep neural networks, and its theoretical properties are also investigated under regularity conditions. Using different synthetic models and a real-world example, we demonstrate the advantage of the proposed algorithm over other alternatives in terms of identifying useful features, avoiding useless ones, and the computation efficiency. Although the algorithm is proposed using standard random forests, it can be widely adapted to other machine learning algorithms, as long as features can be sorted accordingly. 
	
	Key words: Feature importance, Maximum mean discrepancy, Reproducing kernel Hilbert space.
\end{abstract}

\section{Introduction}
\label{sec:intro}
Random forests \citep[RFs]{breiman2001random} are a widely used machine learning algorithm \citep{caruana2006empirical,criminisi2012decision,fernandez2014we}. 
However, their computational efficiency is compromised when they are applied to large-scale datasets with numerous useless features.

Since the landmark of \citet{breiman2001random}, research on RFs has been active in different scientific fields. \citet{NIPS2004_2667} used RFs for structured language learning, and they showed that their method outperformed its competitors in terms of perplexity and error rates. Based on a synthetic dataset generated from a certain reference distribution, \citet{doi:10.1198/106186006X94072} applied RFs to obtain dissimilarities among the original unlabeled data. \citet{NIPS2010_4140} proposed an RFs algorithm to directly estimate the ratio of the proposal and posterior distributions nonparametrically for the Metropolis-Hastings algorithm, and the corresponding theoretical properties were also investigated. Using a Mondrian process, \citet{NIPS2014_5234} proposed a computationally efficient RFs algorithm for online learning, but splits were made independent of the response of interest. Inspired by the local polynomial regression, \citet{pmlr-v70-li17e} proposed a robust RFs algorithm incorporating different loss functions and showed that their method generalized the standard and quantile RFs \citep{meinshausen2006quantile}. \citet{pmlr-v80-haghiri18a} proposed a comparison-based RFs algorithm for the case when the sample was not representative and when it was difficult to measure the distance between instances. \citet{siblini} proposed an algorithm for extreme multi-label learning using a tree-based method and demonstrated that the computation was more efficient than other competitors in parallel. \citet{scornet2015consistency}, \citet{mentch2016quantifying}, and \citet{wager2018estimation} investigated theoretical properties of RFs algorithms. Refer to \citet{criminisi2012decision} and \citet{goel2017random} for a more comprehensive review of RFs.

Albeit it is common to have numerous features in practice, only a limited portion contributes to the response of interest \citep{fan2008sure}. Because existing RFs algorithms do not identify useful features before growing trees, the corresponding computation efficiency is undermined, especially when most features are useless. For example, it is well known that local polynomial regression suffers from the curse of dimensionality; thus, the estimation efficiency of \citet{pmlr-v70-li17e} is questionable. In addition, the online RFs \citep{NIPS2014_5234} may also lead to inefficient estimation when the number of useless features is large, since splits are made based on features regardless of the response of interest. Thus, feature selection is essential for a high dimensional dataset.

Even though RFs  have been actively investigated under different scientific fields, feature selection by RFs does not receive so much attention.
There are two main approaches for feature selection by RFs. One approach is permutation-based, using feature importance \citep{breiman2001random}. \citet{strobl2008conditional} proposed a permutation algorithm to compute conditional feature importance, but they did not provide a general guidance for feature selection. \citet{kursa2010feature} used a new set of shadow features to debias the feature importance, and a feature selection procedure was also proposed based on a ``Z score''; also see \citet{sandri2008bias} for a similar approach. \citet{altmann2010permutation} proposed to permute the response vector to get the ``null importance'', and it was used for feature selection heuristically. \citet{genuer2010variable} proposed a two-step procedure for feature section based on feature importance, and their method worked well for highly correlated features. However, the theoretical properties of existing permutation-based methods have not been rigorously investigated. The other approach is based on the minimum depth proposed by \citet{ishwaran2010high}, and it has been widely applied in survival analysis \citep{Ishwaran2008,twyman2015radiation,benci2016tumor}. The corresponding theoretical properties were investigated by \citet{ishwaran2010high} under strong conditions. For example, the split for a node should always be the median of the values, but such an assumption may undermine the estimation efficiency. 

Herein, we propose a nonparametric feature selection algorithm that incorporates RFs and deep neural networks (NFSRD). Specifically, we adopt nonparametric two-sample tests using deep neural networks \citep{liu2020learning} to select \emph{useful} features, and the corresponding theoretical properties are investigated under regularity conditions. To improve the computational efficiency of the NFSRD, subsampling is adopted. Experiments reveal that the NFSRD outperforms its alternatives in terms of detecting useful features, avoiding useless ones, and the computation efficiency. Another advantage of subsampling is that it saves the computer memory while retaining the desired accuracy for feature selection; refer to Section~\ref{sub: expriment} for details.

The NFSRD differs from existing works in the following aspects. First, we propose the use of nonparametric two-sample tests to select useful features incorporating RFs and deep neural networks, and the corresponding theoretical properties can be rigorously established under regularity conditions. In addition, subsampling is adopted for feature selection; thus, the NFSRD is more computationally efficient than existing RFs-based feature selection algorithms. Moreover, we do not make any strong assumption for splits, as in \citet{ishwaran2010high}. Rather, the only crucial assumption we make is that a limited portion of the features contributes to the response of interest, and such an assumption is also widely adopted for feature selection \citep{fan2008sure}. Besides, the NFSRD adopts a forward feature selection procedure based on the features sorted by their importance, so it can be widely adapted to existing algorithms mentioned in the preceding paragraph and other machine learning algorithms \citep{chen2016xgboost,wager2018estimation}, as long as the features can be sorted accordingly. 

The remainder of this paper is organized as follows. The model setup is introduced in Section~\ref{sec: method}. The detailed algorithm for the NFSRD is presented in Section~\ref{sec: algo}, and the corresponding theoretical properties are also investigated. Simulation studies are presented in Section~\ref{sub: expriment}. Section~\ref{sec: appl} describes an application of the NFSRD to identify useful features based on a superconductivity dataset. The conclusions are provided in Section~\ref{sec: conc rem}.

\section{Model Setup}\label{sec: method}
Let $\bx_i=(x_{i1},\ldots,x_{ip})\in\mathcal{X}$ be a $p$-dimensional feature vector and $y_i\in\mathbb{R}$ be the corresponding response of interest, where $i\in\mathcal{I}_n$, $\mathcal{I}_n=\{1,\ldots,n\}$ is the sample index set of size $n$, and $\mathcal{X}\subset\mathbb{R}^p$. Consider the following regression model:
\begin{equation}
	y_i = f(\bx_i) + \epsilon_i\quad(i\in\mathcal{I}_n),\label{eq: model setup}
\end{equation}
where $f(\bx_i)$ is a smooth function involving $K_0\ll p$ useful features \citep{fan2008sure}, and $\epsilon_i$ is white noise. We are interested in identifying the $K_0$ useful features based on the sample $\mathcal{D}_n=\{(\bx_i,y_i):i\in\mathcal{I}_n\}$.
Without loss of generality, lower cases denote observed data, and upper cases denote the associated random variables. The vectors are column-wise, unless explicitly explained otherwise. 

Before presenting the  NFSRD algorithm, we briefly introduce a bagged-tree learner and RFs. A bagged-tree learner is 
\begin{equation}
	\widehat{\theta}_{TR}(\bx) = \E_*\{t(\bx;\bZ^*_1,\ldots,\bZ^*_n)\},\notag
\end{equation}
where $\bz_i =(\bx_i,y_i)$, and $\{\bZ^*_i:i\in\mathcal{I}_n\}$ is a random sample of size $n$ generated by an empirical distribution $\mbP^*(\bz) = n^{-1}\sum_{i\in\mathcal{I}_n}\ind{\bz =\bz_i}$, $\ind{\bz =\bz_i}=1$ if $\bz=\bz_i$ and 0 otherwise, and $\E_*(\cdot)$ is the expectation with respect to $\mbP^*(\bz)$. For ease of notation, the sample size $n$ is omitted for $\widehat{\theta}_{TR}(\bx)$ and other statistics.

Notably, RFs extend bagged-tree learners by allowing additional randomness within trees to reduce the correlation among them. Specifically, an RFs learner is 
\begin{equation}
	\widehat{\theta}_{RF}(\bx) = \E_*\{t(\bx;\xi,\bZ^*_1,\ldots,\bZ^*_n)\}, \label{eq: random 2}
\end{equation}
where $\xi\sim\Xi$, and $\Xi$ is a pre-specified distribution; see \citet{breiman2001random} for details. A popular choice for $\Xi$ is the random selection of candidate features for the split of each node \citep[\S15.2]{friedman2001elements}. The RFs learner $\widehat{\theta}_{RF}(\bx)$ is a bagged-tree learner $\widehat{\theta}_{TR}(\bx)$ if $\Xi$ is omitted. Bootstrapping is used to approximate $\widehat{\theta}_{RF}(\bx)$ in (\ref{eq: random 2}) by
\begin{equation}
	\widehat{\theta}^{(B)}_{RF}(\bx) = \frac{1}{B}\sum_{b=1}^Bt(\bx;\xi_b,\bz^*_{b1},\ldots,\bz^*_{bn}),\label{eq: RF bootstrap}
\end{equation}
where $B$ is the number of bootstraps, $\xi_b{\sim}\Xi$, and $\bz^*_{bi}$ is a ``realization'' of $\bZ^*_i$ for $b=1,\ldots,B$ and $i\in\mathcal{I}_n$. In practice, a large $B$ is suggested to make the approximation error between (\ref{eq: random 2}) and (\ref{eq: RF bootstrap}) negligible .

\section{NFSRD}\label{sec: algo}
The assumption $K_0\ll p$ validates the feature selection \citep{guyon2003introduction}.
The NFSRD consists of two steps. The first step corresponds to obtaining bias-corrected feature importance (BCFI) by shadow features, and the second step is a forward feature selection based on the \emph{ordered} features by BCFI. Two-sample tests using maximum mean discrepancy and deep neural networks (MMD-D) are conducted sequentially for feature selection in the second step. To improve the computational efficiency, subsampling is applied in both steps. 

Before diving into details, we briefly discuss the intuition for the NFSRD. The proposed method is a forward-stepwise selection algorithm. To avoid including useless features as much as possible, we first order the features based on their ``importance'' for estimating the response of interest. Due to their flexibility for nonparametric modeling, RFs are implemented to order the features by a subset of the instances. Based on the ordered features, forward-stepwise selection is conducted by sequential hypothesis tests. After fitting a full model and a reduced model only retaining the first several important features nonparametrically, the corresponding null hypothesis is that the distributions of the residuals from a full model and a reduced model are identical. If the null hypothesis holds for a certain reduced model, then we treat the involved features as useful and the remaining ones as useless. We still use RFs to train the full and reduced models, and other nonparametric algorithms can be implemented as long as certain consistency results hold; see Supplementary Material and Theorem~1 of \citet{mentch2016quantifying} for details.
\subsection{Bias-Corrected Feature Importance}\label{subsec BCFI}
Feature importance is a rudimentary indicator of the usefulness of features, and it serves as a building block of the NFSRD.

Existing algorithms calculate the feature importance using the entire sample $\mathcal{D}_n$. To achieve better computational efficiency, we propose using subsampling. That is, the feature importance is obtained based on a subsample $\{\bz_i:i\in \mcA_0\}$, where $\mcA_0$ is a subset of $\mcI_n$, and its size is $m_0<n$. The numerical results reveal that $m_0$ should be large to guarantee good performance, and we suggest $m_0\geq 400$ for practical guidance; see Section~\ref{sub: expriment} for details. For example, we can evaluate the feature importance by
\begin{equation}
	F_k = \frac{1}{B}\sum_{t_b}\sum_{j:s_{bj}=k}\Delta(s_{bj},t_b)\quad(k=1,\ldots,p),\label{eq: fi}
\end{equation}
where the first summation is with respect to the $B$ trees, the second summation is with respect the splits made for the $b$th tree, $s_{bj}$ is the feature used to split the $j$th node of $t_b$, $\Delta(s_{bj},t_b) = w_{bj}V_{bj} - w^{(l)}_{bj}V^{(l)}_{bj} - w^{(r)}_{bj}V^{(r)}_{bj}$ is the weighted decrease in variance, $w_{bj} = n_{bj}/m_0$, $n_{bj}$ is the number of instances in the $j$th node of $t_b$, $w^{(l)}_{bj}$ and $w^{(r)}_{bj}$ are the corresponding proportions on the left and right subnodes after splitting the $j$th node by $s_{bj}$, $V_{bj}$ is the sample variance of the response of interest in the $j$th node, and $V^{(l)}_{bj}$ and $V^{(r)}_{bj}$ are variances of the two subnodes. See \citet{breiman2001random} and \citet{sandri2008bias} for details.

However, the feature importance in (\ref{eq: fi}) is unfairly biased toward those with numerous distinct values \citep{white1994bias,louppe2013understanding}. To debias, we incorporate shadow features \citep{sandri2008bias,kursa2010feature}. Specifically, for $i\in\mcA_0$, a shadow feature $\bx^\dagger_i$ is randomly selected from $\{\bx_i:i\in\mcA_0\}$ without replacement. Thereafter, an RF is trained based on the extended data $\{\bz_i^\dagger:i\in\mcA_0\}$ with $\bz_i^\dagger = (\bx_i,\bx^\dagger_i,y_i)$; see \citet{sandri2008bias} for details. The entire procedure is repeated $R$ times, and the BCFI of $X_k$ is calculated as 
\begin{equation}
	I_k = \frac{1}{R}\sum_{r=1}^R(F_{rk} - F_{rk^\dagger})\quad(k=1,\ldots,p),\label{eq: bias-corrected feature}
\end{equation}
where $F_{rk}$ and $F_{rk^\dagger}$ correspond to $X_k$ and $X_k^\dagger$, respectively, and $X_k^\dagger$ is the corresponding shadow feature for $X_k$ for the $r$th repetition; \citet{sandri2008bias} suggest that $R=100$ in practice.
Algorithm~\ref{alg: BCFI} shows the algorithm for BCFI.
\begin{algorithm}[!ht]
	\caption{Bias-Corrected Feature Importance (BCFI)}
	\label{alg: BCFI}
	\footnotesize
	\begin{algorithmic}
		\STATE \textbf{Input:} $\mathcal{D}_n$, $m_0$, and $R$.
		\STATE Select $\mcA_0$ randomly from $\mcI_n$.
		\FOR{$r=1,\ldots,R$}
		\FOR{$i\in\mcA_0$}
		\STATE Generate $\bx_i^\dagger$ from $\{\bx_i:i\in\mcA_0\}$ randomly without replacement.
		\ENDFOR
		\STATE Train an RF using $\{(\bx_i,\bx_i^\dagger,y_i):i\in\mcA_0\}$.
		\STATE Solicit $\{(F_{rk},F_{rk^\dagger}):k=1,\ldots,p\}$.
		\FOR{$k=1,\ldots,p$}
		\STATE Calculate $I_{rk} = F_{rk}-F_{rk^\dagger}$.
		\ENDFOR
		\ENDFOR
		\FOR{$k=1,\ldots,p$}
		\STATE Calculate $I_k = R^{-1}\sum_{r=1}^RI_{rk}$
		\ENDFOR
		
		\STATE \textbf{Output:} $\{I_k:k=1,\ldots,p\}$.
	\end{algorithmic}
\end{algorithm}

\begin{remark}\label{rem: OOB}
	
\end{remark}

We should be cautious about algorithms using feature importance as the \emph{only} criterion for feature selection, especially as no rigorous theoretical properties have been investigated. Thus, in the next subsection, we propose the use of nonparametric two-sample tests for feature selection by ordered features using BCFI.

\begin{remark}\label{rem: other metric}
	Other than the shadow features, another way to debias the  feature importance is to use the out of bag sample (OOB).  \citet{Li2019nips} argued that the OOB-based feature importance outperforms the traditional one in terms of AUC scores.
	In addition, we can also consider the minimum depth \citep{ishwaran2010high} as an importance metric; see Section~\ref{sub: expriment} for details. Thus, those two metrics can  be applied to obtain the feature importance instead. 
\end{remark}
\subsection{Feature Selection by Deep Neural Network}

Ideally, we want to identify a set of features, say $\bX^{(K_0)}$, such that 
$$
Y\perp \bX^{(-K_0)}\mid \bX^{(K_0)},
$$
where $X\perp Y\mid Z$ denotes that $X$ and $Y$ are conditionally independent given $Z$, and $\bX^{(-K_0)}$ contains features other than $\bX^{(K_0)}$. Equivalently, we want to identify $\bx^{(K_0)}$ such that
\begin{equation}
	f(\bx) = f(\bx^{(K_0)}) \quad (\bx\in\mathcal{X}).\label{eq: HT}
\end{equation}

Based on the ordered features by BCFI, a forward feature selection algorithm is proposed by sequentially conducting nonparametric two-sample tests using deep neural networks (FS-D). 
Let the ordered BCFI be $I^{(1)}\geq I^{(2)}\geq\cdots\geq I^{(p)}$ and $x_i^{(k)}$ be the feature corresponding to $I^{(k)}$ for $k=1,\ldots,p$. Consider the following hypothesis testing problem:
\begin{equation}
	H_{0K}: f(\bx) = f(\bx^{(K)})\quad(\bx\in\mathcal{X}),\label{eq: H0 FUNCTIONAKL}
\end{equation}
where $\bx^{(K)}=(x^{(1)},\ldots,x^{(K)})$ contains the first $K$ features with respect to the ordered BCFI, and $K=1,\ldots,p$. 

Although \citet{scornet2015consistency} investigated theoretical properties of an RFs estimator, the convergence rate of $g(\bx)-f(\bx)$ is still an open question, where $g(\bx) = \mbE\{\widehat{\theta}_{RF}(\bx)\}$. Thus, it is difficult to work with (\ref{eq: H0 FUNCTIONAKL}) directly. Instead, we consider $\eta = y - g(\bx)$ and $\eta^{(K)} = y -g(\bx^{(K)})$, where $g(\bx^{(K)}) = \mbE\{\widehat{\theta}_{RF}(\bx^{(K)})\}$. Thus, we have 
\begin{eqnarray}
	\eta^{(K)} &=& y - g(\bx) + [g(\bx) - g(\bx^{(K)})]\notag \\ 
	&=&\eta + [g(\bx) - g(\bx^{(K)})].\notag 
\end{eqnarray}
Instead of $H_{0K}$ in (\ref{eq: H0 FUNCTIONAKL}), we consider
\begin{equation}
	H_{0K}': \mbP = \mbP^{(K)},\label{eq: H0}
\end{equation}
where $\mbP$ and $\mbP^{(K)}$ are the distributions of $\eta$ and $\eta^{(K)}$, respectively. If a feature is useful, it also contributes to $g(\bx)$, and vice versa. Thus, it is valid to use (\ref{eq: H0}) for feature selection.
\begin{remark}
	Because the standard RFs \citep{breiman2001random} is widely implemented in practice, we apply it to obtain BCFI and approximate $g(\bx)$ and $g(\bx^{(K)})$ using (\ref{eq: RF bootstrap}). If we consider an honest tree \citep{wager2018estimation}, we can directly use $g(\bx)$ to approximate $f(\bx)$ according to the consistency result in Theorem~3 of \citet{wager2018estimation}.
\end{remark}
To test $H_{0K}'$ in (\ref{eq: H0}), we adopt a nonparametric two-sample test using the MMD-D \citep{liu2020learning}. The maximum mean discrepancy between $\mbP$ and $\mbP^{(K)}$ is 
\begin{eqnarray}
	\mbox{MMD}(\mbP,\mbP^{(K)};\mathcal{H}_\kappa) &=&\sup_{f\in\mathcal{H}_\kappa;\lVert f\rVert_{\mathcal{H}_\kappa}\leq1}\lvert E\{f(\bX)\}-E\{f(\bX^{(K)})\}\rvert\notag \\ &=&\{E[\kappa(\eta,\eta)] + E[\kappa(\eta^{(K)},\eta^{(K)})]-2 E[\kappa(\eta,\eta^{(K)})]\}^{1/2},\notag \\  \label{eq: test stat}
\end{eqnarray}
where $\kappa:\mathbb{R}\times\mathbb{R}\to\mathbb{R}$ is the kernel for a reproducing kernel Hilbert space (RKHS) $\mathcal{H}_\kappa$, $\lVert f\rVert_{\mathcal{H}_\kappa}$ is the corresponding norm; see Supplementary Material for a brief introduction on RKHS. For a characteristic kernel $\kappa$ \citep{fukumizu2007kernel,gretton2012kernel}, $\mbox{MMD}(\mbP,\mbP^{(K)};\mathcal{H}_\kappa)=0$ is equivalent to $\mbP=\mbP^{(K)}$. Recall that a kernel $\kappa$  is characteristic if the map $\mbQ\to m_{\mbQ}$ is one-to-one, where $m_{\mbQ} = \mbE_{\zeta\sim\mbQ}\kappa(\cdot,\zeta)\in\mathcal{H}_\kappa$ for $\mbQ\in\mathcal{P}_{\mathbb{R}}$, $\mathcal{P}_{\mathbb{R}}$ is the set of probability measures on the measurable space $(\mathbb{R},\mathcal{B})$, $\mathcal{B}$ is the Borel $\sigma$-algebra on $\mathbb{R}$, and $\mbE_{\zeta\sim\mbQ}\kappa(\cdot,\zeta)$ is the expectation of $\kappa(\cdot,\zeta)$ with respect to $\zeta\sim\mbQ$. 

\begin{remark}
	The test statistic is based on (\ref{eq: test stat}), and its intuition is briefly discussed. Evidently, the value $\mbox{MMD}(\mbP,\mbP^{(K)};\mathcal{H})$ is determined by the functional space $\mathcal{H}$. On the one hand, the functional space $\mathcal{H}$ should be sufficiently large to distinguish two different distributions $\mbP$ and $\mbP^{(K)}$ by the supremum of $\lvert E\{f(\bX)\}-E\{f(\bX^{(K)})\}\rvert$ for $f\in\mathcal{H}$. On the other hand, the functional space $\mathcal{H}$ should also be restricted, such that the estimator of $\mbox{MMD}(\mbP,\mbP^{(K)};\mathcal{H})$ converges duly to guarantee good statistical properties. Thus, a unit ball of an RKHS, $\{f\in\mathcal{H}_\kappa;\lVert f\rVert_{\mathcal{H}_\kappa}\leq1\}$, is a good choice for $\mathcal{H}$. See \citet{gretton2012kernel} for details on (\ref{eq: test stat}). 
\end{remark}

If $\{\eta_i:i\in \mcA_1\}$ and $\{\eta_i^{(K)}:i\in \mcA_2\}$ were observed, an estimator of $\mbox{MMD}(\mbP,\mbP^{(K)};\mathcal{H}_\kappa)$ could be obtained by a U-statistic based on two subsamples $\mcA_1$ and $\mcA_2$: \begin{equation}
	\widehat{\mbox{MMD}}_u^2(\widehat{\mbP},\widehat{\mbP}^{(K)};\kappa) = \frac{1}{m_1(m_1-1)}\sum_{i\neq j}H_{ij}^{(K)},\label{eq: MMD}
\end{equation}
where $\widehat{\mbP}(\eta) = m_1^{-1}\sum_{i\in \mcA_1}\ind{\eta=\eta_j}$ is the empirical distribution of $\{\eta_i:i\in \mcA_1\}$, $\widehat{\mbP}^{(K)}$ is the one of $\{\eta_i^{(K)}:i\in \mcA_2\}$, $m_1$ is the size of both $\mcA_1$ and $\mcA_2$, $\mcA_1\cap\mcA_2=\emptyset$, and $H_{ij}^{(K)} = \kappa(\eta_i,\eta_j) + \kappa(\eta_i^{(K)},\eta_j^{(K)}) - \kappa(\eta_i,\eta_j^{(K)}) - \kappa(\eta_i^{(K)},\eta_j)$. The disjoint condition between $\mcA_1$ and $\mcA_2$ guarantees independence between the two error sets. The numerical results reveal that the sample sizes of $\mcA_1$ and $\mcA_2$ should be large, and we suggest that these sizes be larger than 400 for practical guidance ; see Section~\ref{sub: expriment} for details. 

However, $\eta_i$ and $\eta_{i}^{(K)}$ are unavailable; thus, we use $\widehat{\eta}_{ni} = y_i -\widehat{\theta}_{RF}^{(B)}(\bx_i)$ for $i\in \mcA_1$ and $\widehat{\eta}_{ni}^{(K)} = y_i -\widehat{\theta}_{RF}^{(B)}(\bx_i^{(K)})$ $i\in \mcA_2$ instead. The following theorem validates this choice:

\begin{theorem}\label{thm: 1}
	Under mild conditions, in Supplementary Material, $$\widetilde{\mbox{MMD}}_u^2(\widetilde{\mbP},\widetilde{\mbP}^{(K)};\kappa) = \frac{1}{m_1(m_1-1)}\sum_{i\neq j}\widehat{H}_{ij}^{(K)}$$ has the same limiting distribution as $\widehat{\mbox{MMD}}_u^2(\mbP,\mbP^{(K)};k)$ in (\ref{eq: MMD}), where $\widetilde{\mbP}$ and $\widetilde{\mbP}^{(K)}$ are the empirical distributions of $\{\widehat{\eta}_{ni}:i\in \mcA_1\}$ and $\{\widehat{\eta}_{ni}^{(K)}:i\in \mcA_2\}$, $\{\widehat{\eta}_{ni}:i\in \mcA_1\}$ is independent of $\{\widehat{\eta}_{ni}^{(K)}:i\in \mcA_2\}$, and $\widehat{H}_{ij}^{(K)}= \kappa(\widehat{\eta}_i,\widehat{\eta}_j) + \kappa(\widehat{\eta}_i^{(K)},\widehat{\eta}_j^{(K)}) - \kappa(\widehat{\eta}_i,\widehat{\eta}_j^{(K)}) - \kappa(\widehat{\eta}_i^{(K)},\widehat{\eta}_j)$.
\end{theorem} \label{dis: theorem1}
The proof of Theorem~\ref{thm: 1} is relegated to Supplementary Material. The difference between (\ref{eq: MMD}) and the one in Theorem~\ref{thm: 1} is that  the estimated residuals are considered instead. Theorem~\ref{thm: 1} validates that instead of the true residuals, we can use the estimated values, $\{\widehat{\eta}_{ni}:i\in \mcA_1\}$ and $\{\widehat{\eta}_{ni}^{(K)}:i\in \mcA_2\}$, for analysis. Thus, $\widetilde{\mbox{MMD}}_u^2(\widetilde{\mbP},\widetilde{\mbP}^{(K)};\kappa)$ serves as the test statistic for the hypothesis testing problem (\ref{eq: H0}). 

To guarantee independence between $\{\widehat{\eta}_{ni}:i\in\mcA_1\}$ and $\{\widehat{\eta}_{ni}^{(K)}:i\in\mcA_2\}$, we train $\widehat{\theta}_{RF}^{(B)}(\bx)$ and $\widehat{\theta}_{RF}^{(B)}(\bx^{(K)})$ by $\{(\bx_i,y_i):i\in\mcA_3\}$ and $\{(\bx_i^{(K)},y_i):i\in\mcA_4\}$ separately, where $\mcA_3$ and $\mcA_4$ are two subsamples with the same sample size $m_2$, such that $\mcA_1,\ldots,\mcA_4$ are mutually disjoint. Moreover, to avoid correlation, we suggest that $\mcA_1,\ldots,\mcA_4$ are generated from $mathcal{I}_n\setminus \mathcal{A}_0$. For practical guidance, we suggest that $m_2\geq 400$; see Section~\ref{sub: expriment} for details.

\label{dis: deep kernels}
It is hard for simple kernels to distinguish two distributions with complex structures. For example, a translation-invariant Gaussian kernel requires a large sample to distinguish two distributions, since it cannot identify ``direction'' information around each mode in a multivariate setup; see Figure~1 and its discussion of \citet{liu2020learning} for details.   To avoid the drawbacks of traditional parametric kernels, \citet{liu2020learning} proposed obtaining one using  deep neural networks:
$$
\widehat{\kappa}_\omega = \argmax_{\kappa_\omega}\frac{\widetilde{\mbox{MMD}}_u^2(\widehat{\mbP},\widehat{\mbP}^{(K)};\kappa_\omega)}{\widetilde{\sigma}_{1,\lambda}(\widehat{\mbP},\widehat{\mbP}^{(K)};\kappa_\omega)},
$$
where 
$\kappa_\omega = \{(1-\eta)\kappa[\phi_\omega(x),\phi_\omega(y)] + \eta\}q(x,y)$,
$\eta>0$ is predefined, $\phi_\omega(x)$ is a deep neural network with parameter $\omega$ that extracts features, $\kappa(x,y)$ and $q(x,y)$ are Gaussian kernels with lengthscales $\sigma_\phi$ and $\sigma_q$, respectively, and $\widetilde{\sigma}_{1,\lambda}^2(\widehat{\mbP},\widehat{\mbP}^{(K)};\kappa_\omega) = 4m_1^{-3}\sum_{i=1}^{m_1}(\sum_{j=1}^{m_1}H_{ij}^{(K)})^2 - 4m_1^{-4}(\sum_{i=1}^{m_1}\sum_{j=1}^{m_1}H_{ij}^{(K)})^2+\lambda$.

The limiting distributions of the U-statistic $\widetilde{\mbox{MMD}}_u^2(\widehat{\mbP},\widehat{\mbP}^{(K)};\mathcal{H}_\kappa) $ are established below for the null and alternative hypotheses. 
\begin{lemma}\label{lem: asym dis}
	Under the null hypothesis $H_{0K}': \mbP = \mbP^{(K)}$ and regularity conditions in  Supplementary Material, we have 
	\begin{equation}
		m_1\widetilde{\mbox{MMD}}_u^2(\widehat{\mbP},\widehat{\mbP}^{(K)};\kappa_\omega)\to \sum_{i=1}^\infty\zeta_i(Z_i^2-2)\label{eq: 11}
	\end{equation}
	in distribution, where $\zeta_i$ are the eigenvalues satisfying 
	$$ \int\kappa_\omega(\bx,\bz)\Psi_i(\bx)\mbP(\mathrm{d}\bx) = \zeta_i\Psi_i(\bz)$$ for $i=1,2,\ldots,$, $\{\Psi_i(\bx):i=1,2,\ldots\}$ are the eigenfunctions, $Z_i\sim N(0,2)$, and $\mathcal{N}(\mu,\sigma^2)$ is a normal distribution with mean $\mu$ and variance $\sigma^2$.

	Under the alternative hypothesis $H_{aK}': \mbP \neq \mbP^{(K)}$, we have 
	$$
	\sqrt{m_1}[\widehat{\mbox{MMD}}_u^2 - MMD^2]\to \mathcal{N}(0,\sigma^2_1)
	$$
	in the distribution, where $\sigma^2_1 = 4[\mbE(H_{12}H_{13}) - \mbE(H_{12})^2]$.
\end{lemma}
By Theorem~\ref{thm: 1}, we can prove Lemma~\ref{lem: asym dis} in a manner similar to Theorem~12 of \citet{gretton2012kernel}, so its proof is omitted. Instead of deriving the asymptotic distributions in Lemma~\ref{lem: asym dis}, \citet{liu2020learning} suggested permutation for hypothesis testing. Algorithm~\ref{alg: FS} shows the routine of FS-D, and the detailed algorithm for the MMD-D step is relegated to  Supplementary Material.

\begin{algorithm}[!ht]
	\caption{Feature Selection by Deep Neural Network (FS-D)}
	\footnotesize
	\label{alg: FS}
	\begin{algorithmic}
		\STATE \textbf{Input:}$\mathcal{D}_n$, $m_2$, $\{I_k:k=1,\ldots,p\}$, and $\alpha$.
		\STATE Generates mutually disjoint sets $\mcA_1,\ldots,\mcA_4$ from $\mcI_n\setminus\mcA_0$.
		\STATE Train $\widehat{\theta}_{RF}^{(B)}(\bx)$ by $\{(\bx_i,y_i):i\in\mcA_3\}$.
		\STATE Obtain $\{\widehat{\eta}_{ni}:i\in\mcA_1\}$.
		\FOR{$k=1,\ldots,p$}
		\IF{$K=1$}
		\STATE Train KRR by $\{(\bx_i^{(1)},y_i):i\in\mcA_4\}$.
		\STATE Obtain $\{\widehat{\eta}_{ni}^{(K)}:i\in\mcA_2\}$.
		\ELSE
		\STATE Train $\widehat{\theta}_{RF}^{(B)}(\bx^{(K)})$ by $\{(\bx_i^{(K)},y_i):i\in\mcA_3\}$.
		\STATE Obtain $\{\widehat{\eta}_{ni}^{(K)}:i\in\mcA_2\}$.
		\ENDIF
		\STATE Conduct MMD-D with significant level $\alpha$.
		\IF{$H_{0K}'$ is rejected}
		\STATE Continue.
		\ELSE
		\STATE Denote $\widehat{K}_0 = K$.
		\STATE Break the for loop.
		\ENDIF
		\ENDFOR
		\STATE \textbf{Output:} Selected feature set $\mathcal{P}$.
	\end{algorithmic}
	
\end{algorithm}

\begin{remark}
	The basic idea of the NFSRD is to apply nonparametric two-sample tests sequentially based on ordered features. Thus, the NFSRD is also applicable to other machine learning algorithms, such as XGBoost \citep{chen2016xgboost} and the causal tree \citep{wager2018estimation}, as long as the features can be sorted accordingly.
\end{remark}

\section{Simulation}\label{sub: expriment}
In this section, we conduct Monte Carlo simulations to compare the performance of the NFSRD with its alternatives under different model setups. As mentioned in Section~\ref{sec:intro}, there are two main approaches for feature selection by random forests. Thus, we compare the NFSRD with a feature-importance-based algorithm and a minimum-depth-based one.

We use 19 Xeon Cascade Lake (2.5 GHz) CPUs  to train RFs in parallel, and an NVIDIA Tesla T4 GPU is used for FS-D. Five-fold cross validation is conducted to tune the model parameters for RFs; see Supplementary Material for more computational details.

\subsection{Independent Features}
Table~\ref{tab: 1} shows the setups for six synthetic models. Model~1 is linear and is widely used for feature selection \citep{tibshirani1996regression}. Model~2 represents a nonlinear model and only involves one useful feature. Both Model~1 and Model~2 are additive, and Model~3 is more complex and is non-additive. Besides, interaction is involved in Model~3. The difference between Models~1--3 and Models~4--6 is the distribution to generate features. Specifically, features are generated independently from a uniform distribution over $[1,10]$ for Models~1--3, and a \emph{skewed} beta distribution with shape parameter (2,4) is used for Models~4--6. The distribution parameters are chosen such that the features have approximately the same variance in different synthetic models, and the regression parameters are selected such that the signal-to-noise ratio (SNR) ranges from 1 to 3 approximately, where SNR is obtained by $\{\var{[f(\bX)]}/\var{(\epsilon)}\}^{1/2}$, and $\var{(X)}$ is the variance of a random variable $X$. For each model, $\epsilon\sim\mathcal{N}(0,1)$. Furthermore, we consider two feature sizes, including $p=200$ and $p=400$. Although the number of features is large, the useful ones are limited. The sample size is $n=10\,000$. For each synthetic model, we conduct $200$ independent Monte Carlo simulations. 

\if1\inctabfig
{
	\begin{table}[!ht]
		\centering
		\caption{Setups for synthetic models. ``MI'' stands for the model index, and ``Dist'' represents the distribution to generate features. ``Uniform'' corresponds to $X_k\sim\mbox{U}(1,10)$, and ``Beta'' relates to $X_k\sim14.5\mbox{Beta}(2,4)+1$, where $k=1,\ldots,p$. The error term is $\epsilon\sim\mathcal{N}(0,1)$. }\label{tab: 1}
		\footnotesize
		\begin{tabular}{cccc}
			\hline
			MI&Dist& Model Setup&SNR\\
			\hline
			1& Uniform&$Y = 0.3X_1 +0.3X_2 +\epsilon$&1.1\\[0.15cm]
			2& Uniform&$Y = 3\sin( X_1) + \epsilon$&2.1\\[0.15cm]
			3& Uniform&$Y =5 \sin(X_1/10)\sqrt{X_2} +\epsilon$&3.0\\[0.15cm]
			4& Beta&$Y = 0.3X_1 +0.3X_2 +\epsilon$&1.1\\[0.15cm]
			5& Beta&$Y = 3\sin( X_1) + \epsilon$&2.1\\[0.15cm]
			6& Beta&$Y =5 \sin(X_1/10)\sqrt{X_2} +\epsilon$&2.8\\[0.15cm]		
			\hline
		\end{tabular}
		
	\end{table} 
}\fi

We compare the following methods in terms of feature selection under the significance level $\alpha = 0.05$:
\begin{enumerate}
	\item Boruta \citep[BRT]{kursa2010feature}. BRT calculates the feature importance by a set of shadow features, and a ``Z score'' is used for feature selection; see  Supplementary Material for a brief description of BRT. BRT is implemented using the R package \verb|Boruta|.
	\item BRT-N. NFSRD is conducted based on the features sorted by BCFI, as obtained from BRT.
	\item Minimal depth variable selection \citep[MVS] {ishwaran2010high}. MVS selects features using the minimum depth \citep{ishwaran2010high}; see  Supplementary Material for a brief description of MVS. This method is implemented using the R package \verb|randomForestSRC|.
	\item MVS-N. NFSRD is conducted based on the features sorted by the minimum depth obtained from MVS.
\end{enumerate}
Although \verb|VSURF| \citep{genuer2015vsurf} also conducts feature selection by RFs, it is not considered because of its heavy computational burden. For both BRT-N and MVS-N, we consider two scenarios: $m_0=m_1=m_2=200$ and $m_0=m_1=m_2=400$. For a fair comparison, $5m_0$ instances are used for feature selection by BRT and MVS. 

\label{dis: as as }
The four methods are compared in terms of the accuracy of identifying useful features, $\mu_c$, and the average number of useless features included, $n_w$, where 
$\mu_c = (200K_0)^{-1}\sum_{l=1}^{200}\sum_{k=1}^{K_0}\mathds{I}_{lk}$, $n_w =200^{-1}\sum_{l=1}^{200}n_{lw}$, $\mathds{I}_{lk}=1$ is the $k$th useful feature that is correctly identified in the $l$th Monte Carlo simulation, and 0 otherwise, $n_{lw}$ is the number of useless features identified in the $l$th Monte Carlo simulation. The simulation results are summarized in Table~\ref{tab: new2}. When the subsample sizes are small, $\mu_c$ is less than 1 for Model~1, indicating that it is difficult for the two NFSRD algorithms to identify some useful features even for a linear model when the SNR is small. This is because $\widehat{\theta}_{RF}^{(B)}(\bx) - g(\bx)$ or $\widehat{\theta}_{RF}^{(B)}(\bx^{(K)}) - g(\bx^{(K)})$ is not negligible compared with the residual; thus, we cannot test (\ref{eq: H0}) correctly. However, as the SNR increases, both BRT-N and MVS-N identify more useful features on average, even for small subsamples and more complex models. On the one hand, even when $m_0$ is small, BRT and MVS, on the other hand, can correctly select useful features with $5m_0$ instances for different models because $5m_0$ instances are used for feature selection. As $m_1$ and $m_2$ increase, useful features can be correctly identified by BRT-N and MVS-N as well as BRT and MVS for different models. However, BRT-N and MVS-N identify far less useless features compared with BRT and MVS, especially when $m_1$ is large. For example, when $m_1=200$ and $p=400$, BRT and MVS identify 2.17 and 34.88 useless features on average for a linear model, respectively, but the average numbers are only 0.4 and 0.3 for BRT-N and MVS-N. The same conclusion applies to the other setups. As $m_1$ and $m_2$ increase to 400, BRT-N and MVS-N still outperform their alternatives in terms of avoiding useless features.

\if1\inctabfig
{
	\begin{table}[!h]
		\centering
		\caption{Summary of feature selection the four methods, including BRT, BRT-N, MVS, and MVS-N, based on 200 Monte Carlo simulations, where the features are generated independently. ``MI'' stands for the model index. $m_0$, $m_1$ and $m_2$ are the sizes of subsamples for NFSRD algorithms. ``BRT'' represent Boruta, ``MVS'' stands for the minimum-depth variable selection algorithm, and ``BRT-N''  and ``MVS-N'' are NFSRD algorithms with features sorted by BRT and MVS, respectively. $\mu_c$ is proportion that useful features are correctly identified, and $n_w$ is the average length of useless features are included. The best result is highlighted by underline. }\label{tab: new2}
		\footnotesize
		\renewcommand{\arraystretch}{0.75}
		\begin{tabular}{cccrrrrlrrrr}
			\hline
			\multirow{2}{*}{$p$}&\multirow{2}{*}{MI}&\multirow{2}{*}{}&\multicolumn{4}{c}{$m_0=m_1=m_2=200$}&&\multicolumn{4}{c}{$m_0=m_1=m_2=400$}\\
			\cline{4-7} \cline{9-12} 
			&  &  & BRT & BRT-N & MVS &MVS-N &  & BRT & BRT-N & MVS &MVS-N \\ 
			\hline
			\multirow{12}{*}{200} & \multirow{2}{*}{1} & $\mu_c$ & \underline{1.00} & 0.90 & \underline{1.00} & 0.90 &  & \underline{1.00} & 0.98 & \underline{1.00} & 0.98 \\ 
			&  & $n_w$ & 1.75 & 0.38 & 4.91 & \underline{0.34} &  & 1.35 & 0.40 & \underline{0.02} & 0.46 \\ \vspace{-1em}
			&  &  &  &  &  &  &  &  &  &  &  \\ 
			& \multirow{2}{*}{2} & $\mu_c$ &\underline{1.00} &\underline{1.00 }& \underline{1.00} &\underline{1.00} &  & \underline{1.00} & \underline{1.00} &\underline{1.00}&\underline{1.00} \\ 
			&  & $n_w$ & 2.16 & 0.60 & 5.17 & \underline{0.59} &  & 1.42 & 0.54 & \underline{0.12} & 0.49 \\ \vspace{-1em}
			&  &  &  &  &  &  &  &  &  &  &  \\ 
			& \multirow{2}{*}{3} & $\mu_c$ & \underline{1.00} & \underline{1.00} & \underline{1.00} &\underline{1.00} &  & \underline{1.00} & \underline{1.00} & \underline{1.00} & \underline{1.00} \\ 
			&  & $n_w$ & 1.52 & \underline{0.83} & 1.21 & 0.95 &  & 0.98 & 0.82 &\underline{0.00} & 0.88 \\\vspace{-1em} 
			&  &  &  &  &  &  &  &  &  &  &  \\ 
			& \multirow{2}{*}{4} & $\mu_c$ & \underline{1.00} & 0.78 &\underline{1.00} & 0.81 &  & \underline{1.00} & 0.90 & \underline{1.00}& 0.88 \\ 
			&  & $n_w$ & 2.21 &\underline{ 0.26} & 6.83 & 0.32 &  & 1.43 & 0.32 &\underline{0.04} & 0.43 \\ \vspace{-1em}
			&  &  &  &  &  &  &  &  &  &  &  \\ 
			& \multirow{2}{*}{5} & $\mu_c$ & \underline{1.00} & \underline{1.00}& \underline{1.00} & \underline{1.00} &  & \underline{1.00} & \underline{1.00} & \underline{1.00} & \underline{1.00} \\ 
			&  & $n_w$ & 1.98 & \underline{0.66}& 4.42 & 0.71 &  & 1.30 & 0.54 & \underline{0.08} & 0.40 \\ \vspace{-1em}
			&  &  &  &  &  &  &  &  &  &  &  \\ 
			& \multirow{2}{*}{6} & $\mu_c$ & \underline{1.00} & 0.94 & \underline{1.00} & 0.93 &  & \underline{1.00} & \underline{1.00} & \underline{1.00} & \underline{1.00} \\ 
			&  & $n_w$ & 1.57 & 0.78 & 1.61 & \underline{0.75} &  & 1.13 & 0.66 &\underline{0.00} & 0.74 \\ 
			&  &  &  &  &  &  &  &  &  &  &  \\ 
			\multirow{12}{*}{400} & \multirow{2}{*}{1} & $\mu_c$ & \underline{1.00} & 0.86 & \underline{1.00} & 0.85 &  & \underline{1.00} & 0.98 & \underline{1.00} & 0.98 \\ 
			&  & $n_w$ & 2.19 & 0.37 & 23.72 & \underline{0.24} &  & 1.44 & \underline{0.42} & 5.50 & 0.46 \\ \vspace{-1em}
			&  &  &  &  &  &  &  &  &  &  &  \\ 
			& \multirow{2}{*}{2} & $\mu_c$ & \underline{1.00} & \underline{1.00} & \underline{1.00} & \underline{1.00} &  & \underline{1.00} & \underline{1.00} & \underline{1.00} & \underline{1.00} \\ 
			&  & $n_w$ & 2.35 & 0.64 & 19.60 & \underline{0.56} &  & 1.34 & 0.58 & 5.43 & \underline{0.44} \\ \vspace{-1em}
			&  &  &  &  &  &  &  &  &  &  &  \\ 
			& \multirow{2}{*}{3} & $\mu_c$ & \underline{1.00} & 0.99 & \underline{1.00} & 0.99 &  & \underline{1.00} & \underline{1.00} & \underline{1.00} & \underline{1.00} \\ 
			&  & $n_w$ & 1.79 & \underline{1.00} & 3.98 & 1.16 &  & 1.12 & \underline{0.66} & 1.21 & 0.68 \\ \vspace{-1em}
			&  &  &  &  &  &  &  &  &  &  &  \\ 
			& \multirow{2}{*}{4} & $\mu_c$ & \underline{1.00} & 0.90 & \underline{1.00} & 0.90 &  & \underline{1.00} & 0.99 & \underline{1.00} & 0.99 \\ 
			&  & $n_w$ & 2.17 & 0.40 & 34.88 & \underline{0.30} &  & 1.48 & \underline{0.52} & 7.03 & 0.61 \\ \vspace{-1em}
			&  &  &  &  &  &  &  &  &  &  &  \\ 
			& \multirow{2}{*}{5} & $\mu_c$ & \underline{1.00} & \underline{1.00} & \underline{1.00} & \underline{1.00} &  & \underline{1.00} & \underline{1.00} & \underline{1.00} & \underline{1.00} \\ 
			&  & $n_w$ & 2.30 & 0.73 & 28.79 & \underline{0.47} &  & 1.38 & \underline{0.47} & 4.75 & 0.58 \\ \vspace{-1em}
			&  &  &  &  &  &  &  &  &  &  &  \\ 
			& \multirow{2}{*}{6} & $\mu_c$ & \underline{1.00} & 0.87 & \underline{1.00} & 0.88 &  & \underline{1.00} & \underline{1.00} & \underline{1.00} & \underline{1.00} \\ 
			&  & $n_w$ & 1.58 & \underline{0.61} & 17.16 & 0.66 &  & 0.99 & 0.81 & 1.28 & \underline{0.76} \\ 
			\hline
		\end{tabular}
		
	\end{table}
}\fi

We also compare the four methods in terms of computational efficiency, and Figure~\ref{fig: ind} shows boxplots of the computation time based on the 200 Monte Carlo simulations under different setups. When both $m_0$ and $p$ are small, the computational efficiencies of the four methods are comparable, but, as shown in Table~\ref{tab: new2}, BRT and MVS select more useless features than BRT-N and MVS-N. Thus, BRT and MVS are still less favored. As $m_0$ or $p$ increases, the computational efficiency of BRT-N and MVS-N is generally much better, especially when both $m_0$ and $p$ are large. For example, when $m_0=400$ and $p=400$, the computation time required by BRT-N and MVS-N is less than 50 seconds in general, but it is more than 125 seconds and approximately 75 seconds for BRT and MVS, respectively. In addition, it is far less efficient to use BRT and MVS to select features for large datasets with numerous useless features. However, BRT-N and MVS-N can be used to solve this problem. Compared with MVS-N, BRT-N is slightly more computationally efficient in general, but the difference between the two methods is minor.

\if1\inctabfig
{
	\begin{figure}[t]
		\centering
		\includegraphics[width=\textwidth]{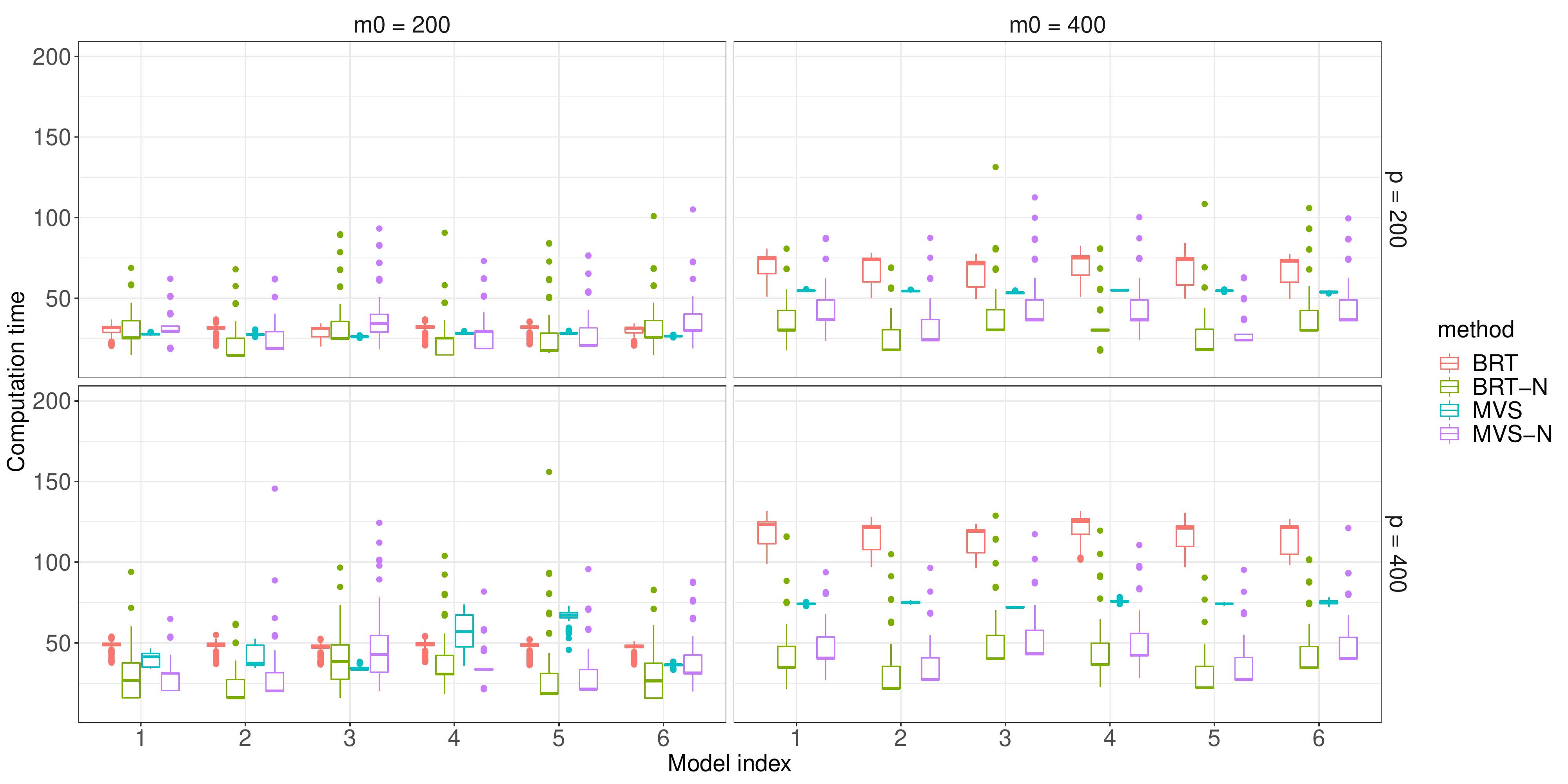}
		\caption{Boxplots for the computation time of different methods based on 200 Monte Carlo simulations, where the features are independently generated. The horizontal segments within the boxplots represent the median computation time. ``BRT'' represent Boruta, ``MVS'' stands for the minimum-depth variable selection algorithm, and ``BRT-N''  and ``MVS-N'' are NFSRD algorithms with features sorted by BRT and MVS, respectively.}\label{fig: ind}
	\end{figure}
}\fi

\begin{remark}
	It is common to have correlated features in practice, and we conduct a simulation study for this case as well. We still consider the same setups in Table~\ref{tab: 1}, but the features are generated differently. First, generate $X_1',\ldots,X_p'$ independently by a uniform or skewed beta distribution, as shown in Table~\ref{tab: 1}. Then, let $X_1=X_1'$ and $X_k = 0.7X_k' + 0.3X_{k-1}'$ for $k=2,\ldots,p$. The features consist of $X_1,\ldots,X_p$, and we still use the six models in Table~\ref{tab: 1} to generate the responses of interest. The simulation results are similar to the aforementioned outcomes, and we relegate them to Supplementary Material.
\end{remark}

\citet{ishwaran2010high} compared the MVS with some commonly used feature selection algorithms, including the adaptive lasso \citep{zou2006adaptive} and the $l_1$-regularized regression model \citep{parkhastie2007}. Their simulation results demonstrated that the MVS outperforms those two in terms of the false discovery rate and the false nondiscovery rate, regardless of the correlation among features. Since the MVS-N is generally more preferable than MVS, we do not consider those two feature selection algorithms in the simulation study.

\section{Application}\label{sec: appl}
The superconductivity dataset \citet{hamidieh2018data} is used to test the performance of the NFSRD. Superconducting materials have wide applications in practice, such as magnetic resonance imaging systems in hospitals and superconducting coils in the Large Hadron Collider at CERN. The accurate prediction of the superconducting critical temperature is important because the corresponding superconductor can only conduct current without resistance at or below this temperature. There are 21\,263 instances in the superconductivity dataset with 81 features extracted for each instance, and the goal is select useful  features for the critical temperature; see \citet{hamidieh2018data} for details about the technical information of the features.

For BRT-N, we consider two significance levels: $\alpha=0.05$ and $\alpha=0.01$, and the size of $\mcA_0,\ldots,\mcA_4$ is $1\,000$. Table~\ref{tab: 2} lists the selected useful features. As the significance level decreases from 0.05 to 0.01, three more features are selected. In addition, all the selected features are among the top 20 most important features by XGBoost \citep{chen2016xgboost}; see Table~5 of \citet{hamidieh2018data} for details. In contrast, BRT blindly identifies all the features to be useful, even for $\alpha=0.05$. Although BRT works reasonably for synthetic models in Section~\ref{sub: expriment}, it fails to identify useful features for superconductivity data. A similar conclusion holds for the two minimum-depth-based methods, and, thus, we omit them.
\if1\inctabfig
{
	\begin{table}[!h]
		\centering
		\caption{The selected features by BRT-N under different significance levels for the superconductivity data. Selected features are indicated by ``$\checkmark$'', and ``Feature'' shows the corresponding names. ``$\alpha$'' is the significant level.}\label{tab: 2}
		\footnotesize
		\begin{tabular}{Hlcc}
			\hline
			VI & \multirow{2}{*}{Feature}&\multicolumn{2}{c}{$\alpha$}\\
			\cline{3-4}
			&& $0.01$&$0.05$\\\hline
			1& \verb|range_ThermalConductivity|&$\checkmark$&$\checkmark$\\
			2&\verb|wtd_std_ThermalConductivity| &$\checkmark$&$\checkmark$ \\
			3&\verb|range_atomic_radius| &$\checkmark$&$\checkmark$\\
			4&\verb|wtd_entropy_atomic_mass| &$\checkmark$&$\checkmark$\\
			5&\verb|wtd_entropy_Valence|&$\checkmark$&$\checkmark$\\
			6&\verb|wtd_mean_Valence|&$\checkmark$&$\checkmark$ \\
			7&\verb|wtd_mean_ThermalConductivity|&$\checkmark$&$\checkmark$\\
			8&\verb|std_ThermalConductivity| &$\checkmark$&$\checkmark$\\
			9&\verb|wtd_gmean_Valence| &$\checkmark$& \\
			10&\verb|wtd_gmean_ThermalConductivity| &$\checkmark$&\\
			11&\verb|wtd_std_ElectronAffinity| &$\checkmark$&\\
			\hline
		\end{tabular}
		
	\end{table}
}\fi

\section{Concluding Remarks}\label{sec: conc rem}
In this study, we propose the NFSRD to identify useful features. Feature selection is conducted by nonparametric two-sample tests using deep neural networks, and the theoretical properties are also investigated. Experiments show that the NFSRD outperforms its alternatives in terms of identifying useful features, avoiding useless ones and the computation efficiency.

Although the NFSRD is proposed using the standard RFs \citep{breiman2001random}, the same idea can be easily adapted to other machine learning methods. In addition, other nonparametric tests can also be used; these include other nonparametric kernel-based methods \citep{scholkopf2018learning} and traditional statistical tests. However, we should pay attention to some widely used tests, such as the $t$-test and the Kolmogorov--Smirnov test, because they may suffer from model misspecification.

\section{Acknowledgements}
The authors thank the editor, associate editor, and two anonymous reviewers for their detailed and constructive comments. 

This work was supported by the National Natural Science Foundation of China (NSFC) (No. 12001109, No. 92046021, No. 11901487, No. 72033002), Science and Technology Commission of Shanghai Municipality grant (No. 20dz1200600) and Fundamental Research Funds for the Central Universities(No.  20720191056).

\bibliographystyle{dcu} 
\bibliography{ref}
\end{document}